\documentclass{article}
\usepackage{ijcai19}

\usepackage{times}
\usepackage{soul}
\usepackage{url}
\usepackage[hidelinks]{hyperref}

\usepackage[utf8]{inputenc}
\usepackage[small]{caption}
\usepackage{graphicx}
\usepackage{amsmath}
\usepackage{booktabs}
\usepackage{csquotes}
\usepackage{float}
\usepackage{wasysym}
\usepackage{xcolor}
\usepackage{subcaption}
\usepackage{ textcomp }
\usepackage{colortbl}

\newcommand{\kge}[1]{KGE#1}

\colorlet{tableheadcolor}{gray!25} 
\colorlet{tablerowcolor}{gray!10} 
\newcommand{\rowcol}{\rowcolor{tablerowcolor}} %

\title{A Comparative Study of Distributional and Symbolic Paradigms \\ for Relational Learning \\ {\normalsize \textcolor{red}{Corrected version: changes in red}}}

\author{
Sebastijan Dumančić$^1$\and
Alberto García-Durán$^2$ \And
Mathias Niepert$^3$
\affiliations
$^1$KU Leuven, Belgium\\
$^2$EPFL, Switzerland \\
$^3$NEC Labs, Germany
\emails
sebastijan.dumancic@cs.kuleuven.be,
alberto.duran@epfl.ch,
mathias.niepert@neclab.eu
}

\hypersetup{draft}
\begin{document}

\maketitle

\begin{abstract}
	\textcolor{red}{This is a corrected version of the original paper, which contained a mistake in evaluation of symbolic models on the KBC tasks. The mistake was identified with the help of Manuel Fink, Christian Meilicke and Melisachew Wudage Checkol (University of Mannheim)}
  Many real-world domains can be expressed as graphs and, more generally, as multi-relational knowledge graphs.
  Though reasoning and learning with knowledge graphs has traditionally been addressed by symbolic approaches such as Statistical relational learning, recent methods in (deep) representation learning have shown promising results for specialised tasks such as knowledge base completion.
  These approaches, also known as distributional, abandon the traditional symbolic paradigm by replacing symbols with vectors in Euclidean space.
  With few exceptions, symbolic and distributional approaches are explored in different communities and little is known about their respective strengths and weaknesses.
  In this work, we compare distributional and symbolic relational learning approaches on various standard relational classification and knowledge base completion tasks.
  Furthermore, we analyse the complexity of the rules used implicitly by these approaches and relate them to the performance of the methods in the comparison.
  The results reveal possible indicators that could help in choosing one approach over the other for particular knowledge graphs.
\end{abstract}

\section{Introduction}

Recent years have created a dichotomy in the field of \textit{Statistical relational learning} (SRL), concerning machine learning with relational data which contains instances and their mutual relationships.
The \textit{symbolic} paradigm \cite{Getoor:2007,Raedt:2016:SRA:3027718}, leveraging the representational and reasoning capacity of first-order logic to compactly represent relational data, has typically dominated the field.
Inspired by the success of (deep) representation learning~\cite{Goodfellow:2016:DL:3086952}, an alternative \textit{distributional} paradigm has emerged, with examples in \textit{Knowledge graph embeddings}~\cite{NickelReview} and \textit{Graph neural networks}~\cite{Hamilton2017RepresentationLO}.
These distributional methods \textit{re-represent} relational data as vectors and/or matrices in the Euclidean space so that the standard feature-based machine learning methods can be used.

The two research directions have largely been developed in isolation and little understanding is currently available on their respective advantages, even though the community has recognised the benefits of both~\cite{DBLP:conf/uai/MinerviniDRR17,NTPs,Manhaeve2018,Evans2018LearningER}.
Moreover, they focus on different tasks and employ different evaluation metrics.
Distributional methods focus on the knowledge base completion (KBC) task measuring the performance through the ranking of proposed entity completions.
Symbolic methods, on the other hand, focus on learning from small relational data with complex forms of logical reasoning and evaluate the performance through the accuracy of predictions.

Contrasting the paradigms upfront, they exhibit a stark contrast on three dimensions: reasoning capabilities, scalability and interpretability.
By relying on logic, the symbolic methods are capable of complex reasoning patterns and are flexible enough to answer any query over a domain (without a need to commit to a predefined target); this is sacrificed with distributional methods~\cite{DBLP:journals/jair/TrouillonGDB19}.
The distributional methods are scalable and can operate on knowledge graphs containing millions of facts, whereas scalability is the major challenge for symbolic methods.
Symbolic methods inherit interpretability from first-order logic, whereas distributional methods are difficult to interpret.
Moreover, distributional methods have difficulties handling unseen instances and, consequently, have to be re-trained every time new data arrives.

This work contributes towards a better understanding of the relative strengths and weaknesses of the aforementioned paradigms.
We focus on the most prominent learning approaches within both paradigms, namely Inductive logic programming and Knowledge graph embeddings, and systematically compare them using standard benchmarks from both communities.
We include both quantitative, in terms of performance, and qualitative analysis, in terms of various data properties, showing that there is no absolute winner amongst the paradigms but data properties (such as neighbour degree and diameter) can help decide which paradigm to use.

\vspace{-5pt}

%
%
\section{Background and Related work}

\textit{Symbolic SRL} methods use first-order logic to represent the data and reason with it.
For instance, a popular symbolic SRL framework Problog~\cite{De-Raedt:2007aa} represents a fact stating that two people (\texttt{marc} and \texttt{eve}) are friends as

\begin{center}
	\texttt{friends(marc,eve).}
\end{center}

Predictive models are expressed as a collection of rules.
For instance, stating that \textit{every person having a friend that smokes is also a smoker} can be expressed in Problog as a \textit{clause}
\begin{center}
	\texttt{smokes(X) :- smokes(Y),friends(X,Y).}
\end{center}

To overcome the disadvantage of deterministic rules concluding only \textit{true} and \textit{false} statements, Problog annotates rules with probabilities to quantify the uncertainty of conclusions.
For instance, one a probabilistic rule
\begin{center}
	0.4 :: \texttt{smokes(X) :- smokes(Y),friends(X,Y).}
\end{center}
states that every person that has a smoking friends has 40 \% of being a smoker.


\textit{Distributional SRL} methods replace symbols (i.e., an entity \texttt{eve} and a relation \texttt{friends}) with vectors and/or matrices in the Euclidean space.
That way the prediction is performed through algebraic manipulations instead of more costly logical inference.
Here we focus on knowledge graph embedding (\kge{}) approaches as the most prominent amongst the existing approaches.
The underlying idea of KGE methods is to associate a score with atoms in a database (i.e., \texttt{friends(marc,eve)}).
Learning then consists of finding the vector representation of instances and their relations by maximising the scores of the atoms in the database and minimising the score of atoms not in the database.
Two prototypical examples are:
\begin{itemize}
	\item \textbf{TransE} \cite{BordesNIPS2013} which interprets relations as translations between instances in the Euclidean space. For each atom \texttt{r(h,t)}, \texttt{r} being a relation and \texttt{h} and \texttt{t} \textit{head} and \textit{tail} instances respectively, the score equals
  				$$ s(r,h,t) = -||\mathbf{e}_h + \mathbf{e}_r - \mathbf{e}_t ||, $$
  				i.e., the vector representation $\mathbf{e}_h$ of a \textit{head} instance translated by the relation vector $\mathbf{e}_r$ should be close to the vector representation $\mathbf{e}_t$ \textit{tail} instance
	\item \textbf{DistMult}  \cite{YangYHGD14a} which focuses on pairwise interactions of \textit{latent} features with the following score
				$$ s(r,h,t) = (\mathbf{e}_h \ocircle \mathbf{e}_r)\mathbf{e}_t^T $$
				where $\ocircle$ is an element-wise product.
\end{itemize}

Checking the validity/truthfulness of any fact comes down to evaluating its score.
The majority of KGE methods proposed so far \cite{EmbeddingsOverview} are a variation on the above scoring functions.

\subsection{Comparing Symbolic and Distributional Methods}

Though limited, several works offer insights into the differences between the two paradigms.
\citeauthor{NickleNIPS2014} [\citeyear{NickleNIPS2014}] and \citeauthor{toutanova2015observed} [\citeyear{toutanova2015observed}] show that including both latent features from \kge{s} and the observable features, in form of random walks over knowledge graphs, in a joint model can greatly increase the performance and reduce the learning complexity.
However, they offer no greater insight into possible reasons.
\citeauthor{pujara:emnlp17} [\citeyear{pujara:emnlp17}] show that \kge{s} have difficulties handling data with a high degree of sparsity and noise -- which is the case with many automatically constructed knowledge graph.
\citeauthor{GrefenstetteTFDS} [\citeyear{GrefenstetteTFDS}] introduces a formal framework for simulating logical reasoning through tensor calculation, which can be seen as a form of embeddings that does not require learning.

The work most related to ours is that of \citeauthor{toutanova2015observed} [\citeyear{toutanova2015observed}], \citeauthor{VigILP2017} [\citeyear{VigILP2017}] and \citeauthor{DBLP:journals/jair/TrouillonGDB19} [\citeyear{DBLP:journals/jair/TrouillonGDB19}].
\citeauthor{VigILP2017} [\citeyear{VigILP2017}] compare symbolic SRL methods with embeddings obtained by the Siamese neural network, and focus on analysing the impact of the available \textit{background knowledge} on the performance.
Their results indicate that \kge{s} might be beneficial when the background knowledge about the task at hand is limited; if such knowledge is available, then the symbolic methods are preferable.
\citeauthor{DBLP:journals/jair/TrouillonGDB19} [\citeyear{DBLP:journals/jair/TrouillonGDB19}] study the which kinds of relational reasoning properties can be captured by distributional models.
Our work presented in this paper differs in a way that it goes beyond quantitative analysis and includes substantial qualitative analysis w.r.t. the dataset properties.

%
%
\section{Aims, Materials and Tasks}

\subsection{Aims}

The main goal of this study is to put the distributional and symbolic relational learning approaches on equal grounds.
Concretely, we focus on the following questions:

\begin{itemize}
  \item[\textbf{Q1}] \textit{How do standard symbolic systems, which can manipulate the relational data directly, compare to the distributional systems, which approximate relational data in Euclidean spaces, on the standard benchmarks from both communities?}
  \item[\textbf{Q2}] \textit{Can we identify data properties which indicate the suitability of individual paradigms?}
\end{itemize}

To answer these questions, we focus exclusively on the classification and completion tasks as they give us a well-defined and clear performance measure, in contrast to the clustering task which is ill-defined. 

We do not perform the runtime comparison as it is rather difficult to assess confidently -- small implementation tricks can make a huge difference in runtimes of the distributional methods.
Likewise, the performance of the symbolic methods depends on the provided language bias -- syntactic instructions on how to construct logical formulas.
This could lead us to wrong conclusions.


\subsection{Materials}

\subsubsection{Datasets}
We focus on standard benchmarks in both communities.
From the symbolic community, we focus on the following datasets: UWCSE, Mutagenesis, Carcinogenesis, Yeast, WebKB, Terrorists and Hepatitis.
The descriptions of datasets can be found in \cite{Dumancic2017}.
From the distributional community, we focus on the FB15k-237 and WN18-RR which are accepted as standard.
The description of the datasets can be found in \cite{dettmers2018conve}.

Relational classification datasets often take the form of a \textit{hypergraph}: an edge (a relationship) can connect more than two instances.
\kge{s} cannot easily handle such datasets as they require relationships to be binary.
To allow \kge{s} to operate on such datasets, we perform the \textit{reification} -- a decomposition of hyperedges to a set of binary edges.

\subsubsection{Symbolic Methods}
Various machine learning methods impose various biases, which plays an important role in machine learning.
To better understand how such biases influence the performance, we experiment with methods from three different families: a relational decision tree \textit{TILDE}~\cite{Blockeel1998285}, a relational version of kernel machines \textit{kFOIL}~\cite{Landwehr:2006:KLS:1597538.1597601} and a kNN with the relational similarity measure of \textit{ReCeNT}~\cite{DumancicMLJ2017}.

The above-mentioned learners belong to a subfield of SRL called Inductive logic programming (ILP)~\cite{Raedt:2008:LRL:1202793}, concerned with building machine learning models in the form of logic programs.
We focus the ILP approaches as learning is more developed than with other symbolic SRL methods which mostly focus on reasoning.

\subsubsection{Distributional Methods}
Due to the sheer number of the existing distributional methods \cite{EmbeddingsOverview}, including a large sample would be infeasible.
Moreover, it is not clear whether a big difference in performance is expected as many methods perform comparably when properly tuned~\cite{DBLP:conf/rep4nlp/KadlecBK17}.
Therefore, we focus on the prototypical and most influential approaches -- TransE, DistMult and ComplEx~\cite{trouillon2016complex}.
Note that ComplEx produces embeddings in the \textit{complex Euclidean space} and that each entity is associated with two embeddings - \textit{real} and \textit{imaginary} one.
In order to create a single embedding out of these two, we concatenate the two embeddings.
Even though TransE is one of the earliest methods,  several recent works show that TransE’s performance is higher than that reported for the “updated” versions of TransE on the same datasets~\cite{DBLP:conf/uai/Garcia-DuranN18}.

Distributional methods are usually trained to assign a high score to all facts in a knowledge base, not to specific \textit{target} facts.
To make sure this training criterion does not put distributional methods at a disadvantage, we use the embeddings as the input data for a classifier learning the predictive model for the pre-specified target.
To match the biases of the symbolic methods, we conjoin the embeddings with a decision tree, SVM and kNN classifiers.
We focus on the \textit{shallow} classifiers, as it would be difficult to disentangle the contribution of latent layers from the quality of embeddings.

As distributional methods focus on the knowledge base completion task, they assume that all instances are given at once and  fill in the missing links in data.
Therefore, handling unseen instances is challenging.
We do not address this issue here, but simply learn the representation of both training and test data at the same time (with labels excluded).
It is, however, worth noting that distributional methods have a certain advantage due to this.

\subsubsection{Dataset Properties}
To better understand conditions under which  each of the paradigms is preferable, we analyse the properties the relational classification datasets and contrast them to the performance of individual methods.
The datasets were transformed into graphs (with reification if necessary) by treating each instance/entity as a node and the relationships as edges, and calculating various graph properties using the \texttt{networkX} package~\cite{Schult08exploringnetwork}.
The full list of the analysed properties is available in Table \ref{tab:informedgraph}.
Some of these properties are only defined over connected graphs (i.e., there is a path between any two nodes in a graph); to ensure this is satisfied, we calculate these properties over the \textit{connected components} of a graph -- a collection of subgraphs in which any two nodes are connected to each other via paths -- and report an average value.

Moreover, the relational classification datasets often distinguish between entities and their attributes.
However, \kge{s} do not make such a distinction but simply treat attribute values as nodes in a graph.
To compensate for this, we analyse two types of graphs: \textit{informed} one which is aware of the attributes, and \textit{uninformed} one which treats attribute values as graph nodes.

Additionally, we include various meta-properties of the datasets.
These reflect properties such as the number of attributes and relations in the datasets, as well as the properties of the components, their size and number.
Given the two types of graphs we can construct from a relational dataset, we include measure aiming at quantifying what is the difference between the two graphs.
To do so, we introduce a measure of \textit{edge reduction} (the proportion of edges that were lost after the conversion to the informed graph) and \textit{degree proportion} (a ratio between the average degree in the informed versus the uninformed graph).

\begin{figure*}[ht]
  \centering
  \includegraphics[width=.85\linewidth]{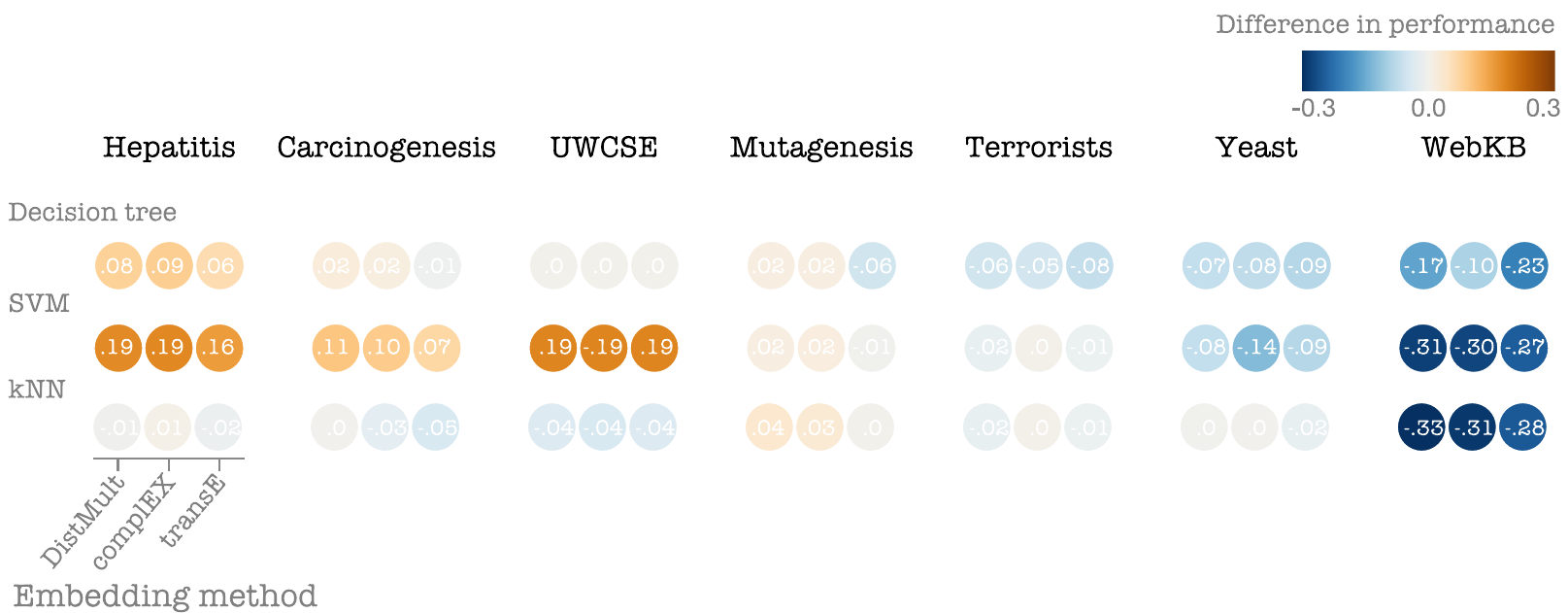}
  \caption{Relative difference in the performance between distributional and symbolic methods (indicated with a colour of a circle representing the embedding method) separates datasets in three groups in the case of decision trees and SVMs: (1) Hepatitis and Carcinogenesis on which the distributional methods work better, (2) UWCSE and Mutagenesis on which the decision is split, and (3) Terrorists, Yeast and WebKB on which the symbolic methods win. The accuracies are calculated in the range [0,1]. No significant difference in performance is observed in the case of kNN. \label{fig:classification}}
  \vspace{-13pt}
\end{figure*}

\subsection{Tasks}

\subsubsection{Relational Classification}
In the relational classification task, certain entities have an associated label and the task is to predict those labels.
We perform standard nested cross-validation (respecting the provided splits) and report the relative performance of the methods in terms of differences in accuracy, $$acc_{distributional} - acc_{symbolic}$$ averaged over individual splits.
The accuracy is reported as a proportion of correct predictions, within the range of [0,1].
The labels were excluded from the data when learning the embeddings and considered only during the training of the classifier.

The embeddings were obtained before learning a classifier.
The dimensions of the embeddings were varied in $\{10, 20, 30, 50, 80, 100\}$; we include smaller dimension because standard relational datasets tend to have a much smaller number of entities than the KBC datasets.
All embeddings were trained to 100 epochs and saved in steps of 20.
We do not use the validation set and metrics such as mean reciprocal rank to select the best hyper-parameters for the embeddings, as they may not be sufficiently correlated with the predictive accuracy; instead, we treat the dimension and the number of epochs for training as additional parameters while training the classifiers as part of the inner cross-validation loop.

\subsubsection{Knowledge Base Completion}
The KBC task corresponds to providing a ranking of missing triplets: given a head/tail entity and a relation, rank the possibilities for the remaining entity.
For the distributional methods, we report the results from \cite{dettmers2018conve} which are considered to be state-of-the-art.
These experiments include several methods: DistMult~\cite{YangYHGD14a}, ComplEx~\cite{trouillon2016complex}, R-CGN~\cite{Schlichtkrull2017ModelingRD} and ConvE~\cite{dettmers2018conve}.

We include TILDE as the representative of the symbolic paradigm, learning a model for each relation.
This was the only symbolic method scaling to the data of this size.
We follow the evaluation procedure as described in \cite{dettmers2018conve} and report the \textit{hits @ K} metric (\textit{is the correct answer among the top K ranked answers}) \cite{BordesNIPS2013}.
In order to calculate the hits@K metric, we need to associate a score with each possible completion to a query.
However, TILDE typically produces binary scores as either true or false.
Thus, to obtain fine-grained scores from TILDE we proceed in the following way: we check which rule triggered each possible completion and take its confidence (i.e., the proportion of correct predictions on the training set) as a score.

\section{Discussion}

\subsection{Relational Classification}

When analysing the results of the relational classification, we are interested in answering the following three questions:
\begin{itemize}
  \item[\textbf{Q1}] How do the two paradigms perform relative to each other?
  \item[\textbf{Q2}] Is there a significant difference in the performance of the embedding approaches?
  \item[\textbf{Q3}] Are there any data properties that correlate with the trend in performance?
\end{itemize}


The results\footnote{Supplementary material: \url{https://arxiv.org/abs/1806.11391} } (Figure \ref{fig:classification} reports the relative difference in performance: positive numbers (orange) indicate that \kge{s} perform better, while the negative numbers (blue) indicate that symbolic methods perform better) indicate that no paradigm is the \textit{absolute} winner: different methods are suitable for different tasks.
However, observing the performance in the case of decision trees and SVMs reveals a pattern dividing the datasets into three groups.
The first group consists of the Hepatitis and Carcinogenesis datasets on which the distributional methods outperform the symbolic approaches.
The second group consists of the Mutagenesis and UWCSE on which the methods either perform similarly or the decision is split (on the UWCSE datasets, symbolic methods perform better with decision trees while distributional methods perform better with SVMs).
The third group contains the Terrorists, Yeast and WebKB datasets on which the symbolic methods perform better.

\begin{table*}[ht]
  \centering
  \resizebox{.95\linewidth}{!}{%
  \begin{tabular}{@{}lrrrrrrr@{}}
    \toprule
    \textbf{Statistics} & \multicolumn{6}{c}{\textbf{Dataset}}\\
    \cmidrule(l){2-8}
    & Hepatitis & Carcinogenesis & UWCSE & Mutagenesis & Terrorists & Yeast & WebKB \\
    \midrule
    average degree  &    2.10(5.64) &   4.62(11.30) &   3.78(8.25) &   3.99(4.58) &   5.02(7.76) &   5.40(12.09) &   2.56(4.17) \\
    \rowcol average neighbor degree & 15.53(1.57) &   28.04(0.00) &   14.25(0.00) &   11.24(0.41) &    4.52(0.20) &    1.57(0.00) &    6.49(1.93)\\
    degree pearson correlation coefficient & -0.66(0.02) &   -0.31(0.00) &     -- &   -0.36(0.00) &     -- &     -- &  -0.18(0.21) \\
    degree assortativity coefficient & -0.66(0.02) &   -0.31(0.00) &     -- &   -0.36(0.00) &     -- &     -- &   -0.18(0.21)\\

    \addlinespace[0.5em]

    average clustering & 0.00(0.00) &    0.60(0.00) &    0.00(0.00) &    0.69(0.00) &    0.28(0.03) &    0.01(0.00) &    0.01(0.00)\\
    square clustering & 0.00(0.00) &    0.76(0.00) &    0.13(0.00) &    0.75(0.00) &    0.30(0.03) &    0.00(0.00) &    0.04(0.02)\\
    \addlinespace[0.5em]

    information centrality & 0.02(0.01) &    0.03(0.00) &    0.30(0.00) &    0.05(0.00) &    0.52(0.03) &    0.92(0.00) &    0.03(0.04)\\
    degree centrality & 0.08(0.03) &    0.09(0.00) &    0.42(0.00) &    0.17(0.01) &    0.68(0.02) &    0.96(0.00) &    0.07(0.10)\\
    closeness centrality & 0.38(0.03) &    0.53(0.00) &    0.57(0.00) &    0.55(0.00) &    0.81(0.01) &    0.96(0.00) &    0.24(0.11)\\
    \addlinespace[0.5em]

    eccentricity & 3.70(0.18) &    1.98(0.00) &    3.55(0.00) &    1.96(0.00) &    1.49(0.03) &    1.19(0.00) &    9.65(2.94)\\
    edge connectivity & 1.00(0.00) &    2.99(0.00) &    1.08(0.00) &   2.00(0.00) &    1.55(0.12) &    1.00(0.00) &    1.00(0.00)\\
    node connectivity & 1.00(0.00) &    1.97(0.00) &    1.08(0.00) &    2.00(0.00) &    1.54(0.13) &    1.00(0.00) &    1.00(0.00)\\
    \rowcol diameter & 4.13(0.18) &    2.00(0.00) &    4.25(0.00) &    2.00(0.00) &    1.65(0.04) &    1.27(0.00) &   13.17(4.35)\\
    radius & 2.23(0.03) &    1.00(0.00) &    2.58(0.00) &    1.00(0.00) &    1.02(0.02) &    1.09(0.00) &    6.75(2.25)\\

    \addlinespace[0.5em]

    maximal clique & 2.00(0.00) &    3.00(0.00) &    2.33(0.00) &    3.00(0.00) &    2.69(0.09) &    2.15(0.00) &    2.67(0.33)\\
    number of maximal cliques & 168.88(19.17) &  109.61(0.00) &  573.67(0.00) &   27.43(1.40) &    4.56(0.19) &  115.67(0.00) & 5213.92(2659.49)\\

    \midrule
    \rowcol number of attributes & 15 & 71 & 23 & 7 & 104 & 1836 & 1967 \\
    number of relations   & 3 & 28 & 5 & 7 & 2 & 1 & 5 \\
    \rowcol edge reduction & .87  & .24  & .22  & .56 & .72  & .73  & .89  \\
    degree reduction & .12 & .78 & .88 & .41 & .44 & .50 & .11 \\

    number of  components & 10.60(1.02) &  340.00(0.00) &   86.00(0.00) &   46.00(0.00) &  221.20(4.62) & 2579.00(0.00) &   51.50(34.09)\\
    \rowcol average component size&160.75(18.37) &   82.83(0.00) &   43.77(0.00) &   26.63(1.19) &    2.73(0.12) &    2.72(0.00) &  723.80(1063.95)\\

    \bottomrule

  \end{tabular}
  }
  \vspace{-6pt}
  \caption{(Meta-)Data properties of the relational classification datasets, with distinguishing the difference between attributes and entities. Among the data properties, only the \textit{average neighbour degree} and the graph \textit{diameter} follow  the performance trend. Among the meta-properties, the \textit{number of attributse}, \textit{edge reduction} and \textit{component size} follow the trend.   \label{tab:informedgraph}}
    \vspace{-15pt}

\end{table*}

Interestingly, when contrasted with the dataset properties of the informed graphs (presented in Table \ref{tab:informedgraph}), only two follow the observed trend: the \textit{average neighbour degree} and the \textit{diameter} of the graph.
We focus the attention of the informed graph, as the statistics of the uninformed graphs did not show any trend related to the performance.
Distributional approaches outperform the symbolic ones on the datasets with a higher neighbour degree (Hepatitis and Carcinogenesis, a split decision of UWCSE and Mutagenesis).
This property, calculating \textit{how many neighbours my neighbours have}, indicates the density of interactions among the nodes in a connected component: higher values mean there are more interactions between the nodes.
It suggests that, for the relational classification, the distributional approaches outperform the symbolic ones on densely connected graphs.
The diameter of a graph is the \textit{longest shortest path between any pair of nodes}, i.e., it estimates whether the connected components are \textit{compact} (nodes are close to each other) or \textit{spread out} (many nodes are far apart).
According to this measure, the symbolic methods perform better in the extremal part of the range (covering the lower and the upper part of the range), while distributional methods prefer the middle range.

Several meta-properties correlate with the performance trend.
Firstly, the datasets on which the symbolic methods perform better have a substantially larger number of attributes than the ones on which the distributional methods win.
Secondly, the average size of the components follows the trend of the diameter: the symbolic methods perform better at the extremal part of the range.
Thirdly, according to the \textit{edge reduction} measure, the symbolic methods perform better when the majority of edges in the uninformed graph actually belongs to the attribute-value assignment, i.e., the edge reduction values are higher.
The exception to this is the Hepatitis dataset,  which still satisfies the observation that the symbolic methods work better on the datasets with more attributes.

Though these measures indicate correlation and might not be a definite way to make a decision, they offer a simple test that indicates a preference for a certain paradigm.
All of the indicative properties can be calculated efficiently using only the data.
The reliance of the symbolic methods on search procedures explains why they underperform on the datasets with  higher neighbour degree: increasing this value enlarges the search space and the symbolic methods, which rely on the local search procedures, might be stuck in the local optima.
Distributional methods might be better suited for such scenarios as they leverage all available information by design.

The observed pattern, however, disappears with the kNN classifier.
The distributional methods gain the advantage on the Mutagenesis datasets, but the symbolic approaches are favoured on the remaining datasets.
However, a general trend is that the differences in performances are much less pronounced compared to the results with decision trees and SVMs, except on the WebKB dataset.
These results indicate that the embedding methods work as well as the manually designed methods for estimating the similarity of relational objects and, therefore, form a viable alternative.

Comparing the performances of the embedding methods indicates that there is no significant difference between them.
This is an interesting observation in itself as it indicates that, when it comes to relational classification, the choice of the embeddings methods matters less than the choice of the classifier.

\subsection{Knowledge Base Completion}

\begin{table}
	\centering
  \resizebox{\linewidth}{!}{%
		\begin{tabular}{@{}lrrrrrr@{}}
		\toprule
						& \multicolumn{3}{c}{\textbf{FB15-237}} & \multicolumn{3}{c}{\textbf{WN18-RR}} \\
			\cmidrule(lr){2-4} \cmidrule{5-7}

						          & Hits@1  & Hits@3    & Hits@10 	      & Hits@1   & Hits@3    & Hits@10 \\
			\midrule

		\textbf{TILDE} 	  & \textcolor{red}{.12}		& \textcolor{red}{.27}	& \textcolor{red}{.28} & \textcolor{red}{.16}	  & \textcolor{red}{.16}			   & \textcolor{red}{.16} \\
		\midrule
		\textbf{ConvE}	  &	.327		&	.356			&	.501				    & .40			&	.44				 &	.52		\\
		\textbf{Complex}  &	.158		&	.275			&	.428				    & .41			&	.46				 &	.51		\\
		\textbf{DistMult} &	.155		&	.263			&	.419				    & .39			&	.44				 &	.49		\\
		\textbf{R-GCN}	  & .153		&	.258			&	.417				    &	--			&		--		   &		--	\\
		\bottomrule

		\end{tabular}
	}
	\vspace{-7pt}
	\caption{TILDE obtains higher scores on the KBC task, compared to the common distributional approaches \vspace{-5pt}}\label{tab:kbcres}
	
	\vspace{-10pt}

\end{table}

\begin{figure*}[ht]
  \begin{subfigure}{0.33\linewidth}
    \centering
    \includegraphics[width=.7\linewidth]{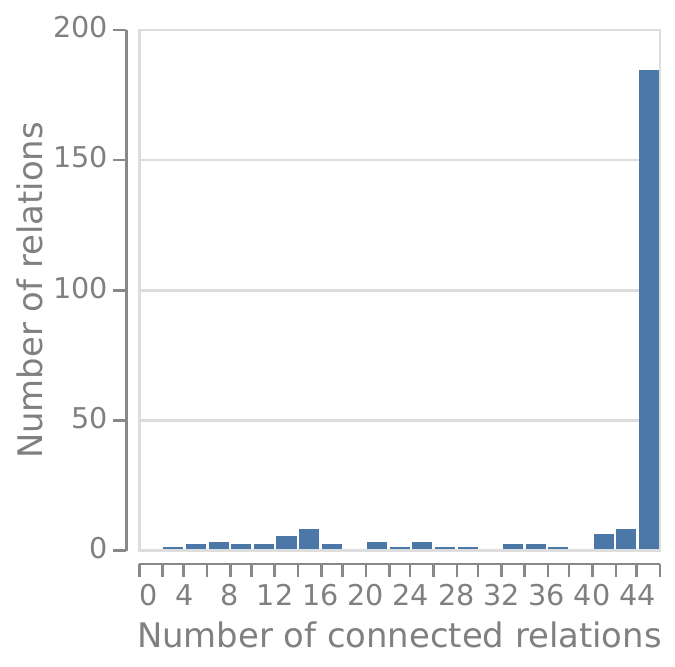}
    \caption{Number of connected relations\label{fig:numrelationsentity}}
    \vspace{-5pt}
  \end{subfigure}
  ~
  \begin{subfigure}{0.33\linewidth}
    \centering
    \includegraphics[width=.7\linewidth]{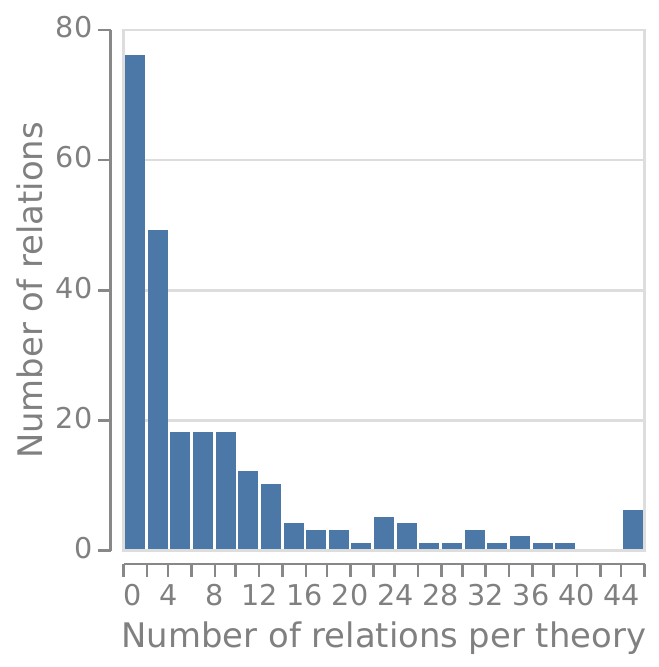}
    \caption{Number of relations per theory\label{fig:numrelationsrule}}
    \vspace{-5pt}
  \end{subfigure}
  ~
  \begin{subfigure}{0.33\linewidth}
    \centering
    \includegraphics[width=.7\linewidth]{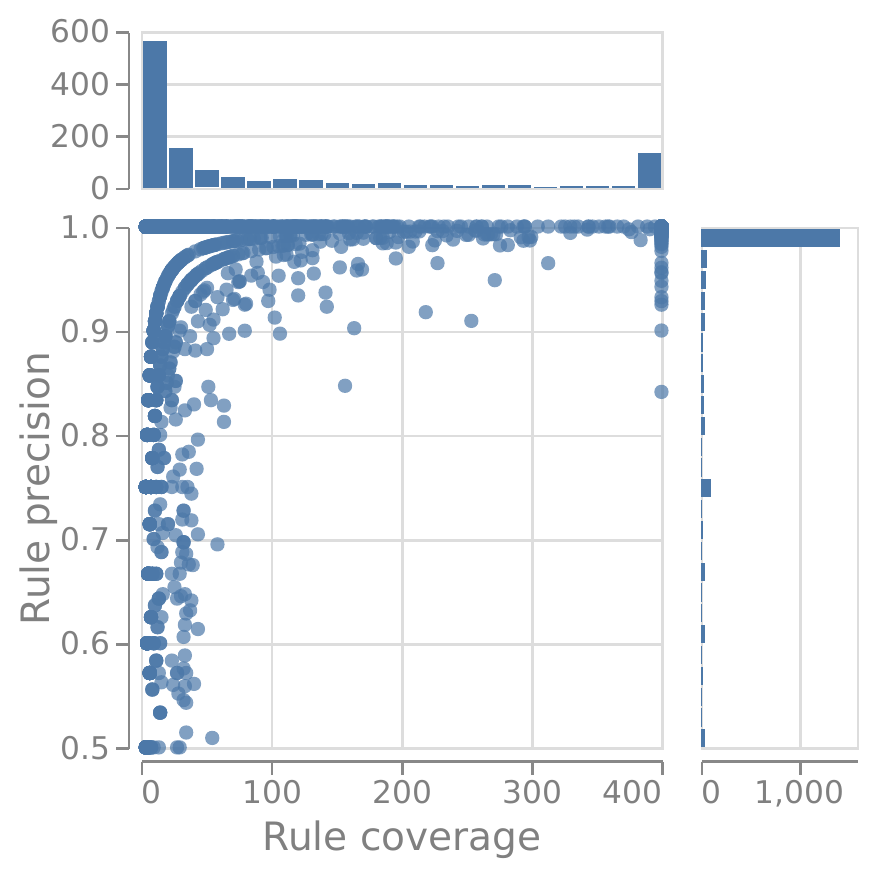}
    \caption{Rule precision vs coverage\label{fig:precvscov}}
    \vspace{-5pt}
  \end{subfigure}
  \caption{The histograms summarise the properties the data in terms of the connected relations (\ref{fig:numrelationsentity}) and the properties of the rules as a number of distinct relations TILDE rules use to predict the existence of a specific relation (\ref{fig:numrelationsrule}). We see these distributions are very different, indicating that only a small proportion of data is useful for the predictions. The third figure summarised the rule properties in terms of their confidence and coverage.}

    \vspace{-15pt}
\end{figure*}

\textcolor{red}{The results on the KBC task (Table \ref{tab:kbcres}) indicate that the symbolic methods fall behind the distributional methods.
	We provide the the analysis of failure modes of the symbolic methods in the appendix.}


To understand better  why is this the case, we analyse the rules learned by TILDE on the FB15k dataset.
Specifically, we inspect their complexity -- the number of relations the rules are composed of, and the number of rules per relation -- and their effectiveness -- estimated by the precision (out of all predictions by a rule, how many of them are correct) and coverage (how many triplets a rule covers).
We discuss two  observations.

First, it is interesting to contrast the number of \textit{connected relations} in data -- given a relation, \textit{how many other relations share an entity with that relation} (Figure \ref{fig:numrelationsentity}) -- and the number of distinct relations TILDE uses to predict the existence of a triplet for a specific relation (Figure \ref{fig:numrelationsrule}).
The former tells us how much information we have about a relation, while the latter estimates how much information we need to predict the existence of the relation between two entities.
Interestingly, we observe a stark contrast between the two distributions: whereas most relations are connected to more than 40 other relations, the majority of TILDE rules for a specific relation contain up to only four distinct relations.
That is, out of 40 relations an entity participates in, only a small fraction of those is useful to predict the existence of other relationships.
Therefore, useful information is sparse.

Second, the extracted rules have a high precision (Figure \ref{fig:precvscov}, right), i.e., how many of the predictions made by a rule are correct (meaning that they exist in the ground truth), despite their simplicity: for the vast majority of rules, more than 90\% of predictions are correct.
If we contrast that with the coverage of the rules, i.e., how many triplets a rule declares as true, we see that the rules with the highest precision are not only the ones with small coverage; instead, the rules with high coverage are spread through the coverage spectrum, including the rule covering hundreds and thousands of triplets (the last bin includes rules with a coverage higher than 400).
Moreover, the rules with the lowest precision have a small coverage.

These two observations suggest that a big advantage of TILDE is that it can be selective about which information it uses to predict the existence of relations between pairs of entities.
The knowledge graph embeddings do not have this ability and, by design, use all available information for link prediction.

A disadvantage of the symbolic methods are the longer training times on the KBC tasks.
However, the rules have to be extracted only once and can be applied every time a new data arrives.
Distributional methods, on contrary, have to be re-trained every time a new entity arrives, as they cannot handle unseen data introducing new entities whose embeddings cannot be calculated.

\section{Conclusion}

Many problems nowadays are naturally expressed in the form of relational and graph structured data.
This includes social and protein interaction networks, biological data, knowledge graphs and many more.
Two main machine learning paradigms for analysing such data -- the symbolic paradigm that relies on first-order logic to represent and manipulate relational data, and the distributional paradigm which re-represents relational data in vectorised Euclidean space -- have mostly been studied in isolation.
This work is among the first, to the best of our knowledge, that systematically compares these two paradigms on the standard tasks from both domains -- relational classification and knowledge base completion.

We draw several conclusions from the experimental analysis.
First, there is no absolute winner among the paradigms, but (meta-)data properties of the relational classification datasets can help to decide which paradigm to prefer.
Second, estimating the similarity of relational objects is an open question in the symbolic community and the distributional SRL methods constitute a viable alternative to manually designed similarities.
Third, the symbolic methods fall behind the distributional ones on the task of knowledge base completion.

This work is not meant as the criticism of any of the considered approaches, but rather a step towards better understanding and integration.
We hope this work inspires new research directions focused on the combination of the paradigms which the community has started to explore  \cite{DBLP:conf/uai/MinerviniDRR17,Schlichtkrull2017ModelingRD}, but many questions remain open.
Most importantly, better methods quantifying reasoning abilities of distributional methods are needed.

\vspace{-2pt}

\section*{Acknowledgements}
\vspace{-2pt}

The authors wish to thank Tim Dettmers and Jonas Schouterden for their help with the experiments. This work was partially funded by the VLAIO-SBO project HYMOP (150033) and FWO (K204818N).
\textcolor{red}{The authors are thankful to Manuel Fink, Christian Meilicke and Melisachew Wudage Checkol (University of Mannheim) for helping identify the bug in the experiments.}

\vspace{-1pt}

\bibliographystyle{named}
\bibliography{starai}

\newpage

\appendix

\section{Failure modes of symbolic methods on KBC tasks}

There are several factors that contributed to the initial overly-optimistic results by applying TILDE on the KBC datasets.
They come from the un-anticipated interaction between the evaluation metric and certain properties of TILDE.

\subsection{Evaluation problem}

In the initial experiments, we have followed the procedure described in the work of \citeauthor{dettmers2018conve} (\citeyear{dettmers2018conve}), which calculates the rank of the $i$-th test triplet as
$$ \text{rank}^s_i = 1 + \sum_{\tilde{x_i} \in \mathcal{C}^s(x_i) \setminus \mathcal{G}} I [ \psi(x_i) < \psi(\tilde{x_i})],$$

where $\tilde{x_i}$ is a corrupted triplet and $\psi$ is the scoring function.
Calculating the rank this way \textit{implicitly assumes} that every triplet has a different rank (or at least that there are very few triplets with the same rank).
Consequently, this mean that in the case of tie, the ground truth triplet is considered to be ranked first amongst the equals.
To compensate for that, we use the \textit{expected rank} (proposed by Manuel Fink) defined as

\begin{align*}
	\text{rank}^s_i = & 1 + 0.5\sum_{\tilde{x_i} \in \mathcal{C}^s(x_i) \setminus \mathcal{G}} I [ \psi(x_i) < \psi(\tilde{x_i})] \\
					  & + 0.5\sum_{\tilde{x_i} \in \mathcal{C}^s(x_i) \setminus \mathcal{G}} I [ \psi(x_i) \leq \psi(\tilde{x_i})],
\end{align*}

i.e., the average of (1) ranking the ground truth triplet as the first amongst  equals and (2) ranking it as the last amongst equals.

\textbf{You can read more about this problem and its spread in \cite{EmbProblem} }

\subsection{Issues with TILDE}

The above outlined issue is not necessarily a problem for symbolic approaches, even though they do produce a discrete set of confidence scores (in the case of TILDE, one per leaf of the tree) as long as the number of false positives is low.
We aimed to verify this with the experiments related to Figure \ref{fig:precvscov} by checking the coverage and precision of the rules.
However, these were checked only on the samples of data (i.e., including only a sample of true negatives) and might give a wrong picture.

The problem with TILDE comes from the way it evaluates the performance.
Instead of using ranking as the distributional approaches, the evaluation criterion TILDE uses is \textit{logical entailment}: how many of the provided example a current model covers?
As the coverage is the only thing that matters (and the number of false positives that would be generated that way does not matter), TILDE can construct rules such as

\begin{center}
	\texttt{relationA(X,Y) :- relationB(X,Z), relationC(Z,W).} \\
	
	\texttt{relationA(X,Y) :- relationB(X,W), relationC(Y,Z).}
\end{center}

Both of these rules would generate a lot of false positives because either (1) one of the arguments of the head atom does matter, or (2) the head arguments are disconnected.
TILDE does not offer an effective way to prevent this from happening.
The number of such rules can be minimised by providing a lot of negative samples and hoping that such cases would be among the negative examples.
Besides that, we have filtered out every rule where either one of the head arguments does not matter or the head arguments are disconnected.

We believe  this clearly outlines two important issues for the ILP community: (1) is entailment truly sufficient or should other objective function be considered?, and (2) what are the sensible classes of languages bias such that above-outlined cases can be avoided?

\section{Hyper-parameters}

This section provides the details on the hyper-parameter values we optimised during the experiments.

\subsection{Propositional classifiers}

We tune the following parameters of the \texttt{sci-kit learn} learners:

\paragraph{\textbf{k Nearest Neighbours}}
\begin{itemize}
	\item \textit{k} - the number of neighbours: $[3, 5, 7, 9, 11, 13, 15]$
	\item \textit{weights} of the neighbours: $["uniform", "distance"]$
\end{itemize}

\paragraph{\textbf{Decision tree}}
\begin{itemize}
	\item training \textbf{criterion}: $["gini", "entropy"]$
	\item \textit{maximal depth} of the tree: $[4, 8, 12, 16, 24, 48]$
	\item \textit{minimal number of samples in the leaf} of a tree: $[2, 4, 6, 8]$
	\item \textit{minimal impurity decrease}: $[0.0, 0.05, 0.1, 0.15]$
\end{itemize}

\paragraph{\textbf{SVMs}}
\begin{itemize}
	\item \textit{kernel} type: $["rbf", "linear", "poly"]$
	\item \textit{degree} of the polynomial kernel: $[3, 5, 7, 9]$
	\item \textit{C} parameter: $[0.1, 1.0, 10.0, 100.0]$
	\item \textit{gamma} parameter: $[0.01, 0.1, 0.5, 0.05, 0.2, 0.3, 0.6, 0.7, 1]$
\end{itemize}

\subsection{Relational classifiers}

We tune the following parameters of the relational classifiers:

\paragraph{\textbf{TILDE}}
\begin{itemize}
	\item \textit{learning heuristic}: $["gain", "gainratio"]$
	\item \textit{minimal accuracy of rules}: $[0.75, 0.8, 0.9, 1.0]$
	\item \textit{minimal examples (cases) in the leaves}: $[2, 4, 6, 8, 10, 12, 15]$
\end{itemize}

\paragraph{\textbf{Relational kNN}}
\begin{itemize}
	\item \textit{weights} of the ReCeNT similarity measure: the individual weights come from the range $\{0.0, 1.0\}$ by increments of $0.05$ and the constraint $\sum_{i} = 1.0$
	\item \item \textit{k} - the number of neighbours: $[3, 5, 7, 9, 11, 13, 15]$
\end{itemize}

\paragraph{\textbf{kFOIL}}
\begin{itemize}
	\item \textit{maximal number of clauses}: $[25, 100, 1000, 10000]$
	\item \textit{maximal number of literals in a clause}: $[3, 5, 7, 9]$
	\item \textit{beam size}: $[1, 2, 3, 4, 5, 6, 7]$
	\item \textit{kernel type} t: $ [0, 1, 2, 3]$
	\item \textit{C} parameter (termed $g$ in kFOIL): $[0.1, 1.0, 10.0, 100.0]$
	\item \textit{s} parameter of the polynomial kernel: $[0.01, 0.05, 0.1, 0.5, 1.0, 2.0, 10.0]$
	\item \textit{r} parameter of the polynomial kernel: $[0.5, 2.0, 1.0, 0.1, 10.0, 0.01]$
	\item \textit{used biased hyperplane}: $[true, false]$
	\item \textit{Normalise kernel after iteration}: $[true, false]$
	\item \textit{move points according to centre of mass in feature space}: $[true, false]$
	\item \textit{trade-off between training error and margin}: $[0.1, 0.25, 1.0]$
\end{itemize}

\section{Experimental details}

\subsection{Code repository}

The code developed for the experiments is available in the following repository: \texttt{\url{https://bitbucket.org/sdumancic/embeddingsmeetilp}}

\subsection{Further elaboration on selected baselines}

The selection of symbolic methods included in the experiments might seem limited initial, especially as we ignore existing methods that combine logical and probabilistic reasoning, such as Markov Logic Networks (MLN), Probabilistic Soft Logic and ProPPR.
We agree all these frameworks are an important part of SRL. However, we don’t include them for two reasons. 
First, learning logic programs from data is much better developed within ILP (Our interpretation of SRL is ILP + probabilities). We have experimented with the MLN learners but have encountered 2 issues: (1) some of the datasets were too big for MLNs, and (2) MLN learners cannot learn clauses with constants, which are essential for the majority of the tasks. Often these MLN learners would learn nothing unless a dataset is somehow reduced (for instance, selecting only the top 100 words in the WebKB dataset). ProPPR is likewise unable to learn clauses with constants, and we are unaware of any PSL structure learner. Regarding the RDNs, we are aware of the BoostSRL implementation, but it would be difficult to marginalise out the effect of boosting. 
Regarding other commonly used ILP techniques such as Aleph and Metagol, we had difficult time to make them work on all considered datasets. Regarding Metagol, it was difficult to provide a generally applicable set of templates beyond dyadic template discussed in the paper which were not sufficient. Regarding Aleph, we could not make it work for all datasets, resulting in many models that would simply predict majority class.

Second, we believe that keeping the probabilistic aspect separate allows us to make a clearer comparison. Distributional approaches aim at representing the graph-structured data in a vectorised format. Therefore, they try to capture relevant structure, and the uncertainty scores they provide reflect the certainty of the model that a link exists. This is related to capturing the relational information, but not a probabilistic aspect of it. In that sense, they are more related to the ILP methods. Moreover, distributional methods do not provide a probabilistic model of a domain and are closer to approximate data lookup techniques.

\subsection{Language specification for symbolic methods}

The specification of the language bias for the symbolic methods can be found here: \texttt{\url{https://bitbucket.org/sdumancic/embeddingsmeetilp/src/master/experiments/workflow_case_classification.py}}

\subsection{Absolute performance of individual methods}

The performance of individual methods on the relational classification datasets, in terms of accuracy (= a proportion of correct predictions), is reported in Table \ref{tab:rawperformance}.

\subsection{Data properties}

The properties of the uninformed graph are reported in Table \ref{tab:uninformedgraph}.

\begin{table*}[ht]
  \centering
  \caption{Dataset properties, uninformed graph \label{tab:uninformedgraph}}
  \resizebox{.95\linewidth}{!}{%
  \begin{tabular}{@{}lrrrrrr@{}}
    \toprule
    \textbf{Statistics} & \multicolumn{6}{c}{\textbf{Dataset}}\\
    \cmidrule{2-7}
    & Hepatitis & Carcinogenesis & Mutagenesis & Terrorists & Yeast & WebKB \\
    \midrule
    degree assortativity coefficient&-0.49(0.01) &      -0.11(0.00) &      -0.33(0.00) &      -- &      -- &        -0.17(0.01)\\
    graph clique number&2.00(0.00) &       3.00(0.00) &       3.00(0.00) &     2.81(2.54) &     2.63(2.92) &       3.00(0.00)\\
    average neighbor degree&324.19(170.83) &   248.59(361.31) &   195.23(72.76) &   32.70(24.55) &  399.70(519.56) &   206.92(239.57)\\
    radius&4.00(0.00) &       4.00(0.00) &       3.00(0.00) &     1.26(0.94) &     1.53(1.39) &       4.50(0.50)\\
    graph number of cliques&14213.40(724.85) & 55647.00(0.00) &    4755.60(218.39) & 196.93(696.61) &2051.53(8477.26) &81606.25(16861.81)\\
    degree pearson correlation coefficient&-0.49(0.01) &       -0.11(0.00) &     -0.33(0.00) &      -- &       -- &       -0.17(0.01)\\
    edge connectivity&1.00(0.00) &        1.00(0.00) &      1.00(0.00) &     1.12(0.36) &     1.05(0.22) &       1.00(0.00)\\
    node connectivity&1.00(0.00) &        1.00(0.00) &      1.00(0.00) &     1.12(0.36) &     1.05(0.22) &       1.00(0.00)\\
    eccentricity&5.11(0.32) &        5.34(0.49) &      4.12(0.34) &     5.58(1.27) &     7.27(1.91) &       6.41(0.74)\\
    average clustering&0.00(0.00) &        0.49(0.02) &      0.22(0.02) &     0.11(0.27) &     0.01(0.02) &       0.01(0.00)\\
    information centrality&-- &          -- &         -- &      0.03(0.15) &       -- &        --\\
    degree centrality&0.01(0.03) &        0.00(0.00) &      0.01(0.03) &     0.06(0.18) &     0.01(0.05) &       0.00(0.01)\\
    diameter&6.40(0.49) &        8.00(0.00) &      6.00(0.00) &     1.87(1.68) &     2.68(2.34) &       8.50(0.50)\\
    square clustering&0.01(0.02) &        0.02(0.02) &      0.02(0.04) &     0.22(0.34) &     0.04(0.17) &       0.02(0.08)\\
    closeness centrality&0.37(0.05) &        0.26(0.02) &      0.40(0.03) &     0.34(0.14) &     0.33(0.08) &       0.26(0.03)\\
    \bottomrule

  \end{tabular}
  }

\end{table*}

\begin{table*}
	\centering
	\caption{Absolute performances, in terms of accuracy, of all classifiers in the analysis\label{tab:rawperformance}}
	\resizebox{.9\linewidth}{!}{%
		\begin{tabular}{@{}lrrrrrrr@{}}
			\toprule
		 		& \multicolumn{7}{c}{\textbf{Datasets}} \\
		 		\cmidrule{2-8}
		 		& Hepatitis & Carcinogenesis & UWCSE & Mutagenesis & Terrorists & Yeast & WebKB \\
		 	\midrule
		 	\textbf{Decision trees} & \multicolumn{7}{c}{} \\
		 	\quad TILDE & .81 & .60 & .95 & .75 & .83 & .87 & .74 \\
		 	\quad DistMult & .90 & .63 & .95 & .77 & .76 & .89 & .56 \\
		 	\quad TransE & .88 & .59 & .95 & .69 & .76 & .88 & .51\\
		 	\quad ComplEx & .90 & .62 & .95 & .77 & .74 & .89 & .64\\
		 	\addlinespace[1em]
 			
 			\textbf{SVM} & \multicolumn{7}{c}{} \\
		 	\quad kFOIL & .8 & .5 & .75 & .77 & .77 & .86 & .89 \\
		 	\quad DistMult & .95 & .61 & .95 & .79 & .75 & .78 & .58 \\
		 	\quad TransE & .93 & .58 & .95 & .77 & .75 & .77  & .62 \\
		 	\quad ComplEx & .95 & .60 & .95 & .79 & .77 & .72 & .59\\
		 	\addlinespace[1em]
		 	
		 	\textbf{kNN} & \multicolumn{7}{c}{} \\
		 	\quad ReCeNT & .81 & .64 & .95 & .85 & .85 & .86 & .89 \\
		 	\quad DistMult & .80 & .64 & .91 & .81 & .83 & .86 & .56 \\
		 	\quad TransE & .79 & .59 & .91 & .85 & .85 & .84 & .61\\
		 	\quad ComplEx & .80 & .61 & .91 & .82 & .83 & .86 & .57 \\
			\bottomrule
		\end{tabular}
	}
\end{table*}


\end{document}